\documentclass[runningheads]{llncs}
\usepackage{graphicx}
\usepackage{amsfonts}
\usepackage{amsmath}
\usepackage{algorithmic,algorithm} 
\usepackage{url}
\usepackage{multicol}
\usepackage{ulem}
\usepackage{color}
\usepackage{cite}
\usepackage{subcaption}
\usepackage[title]{appendix}

\newcommand{\Aclause}[0]{\textsc{a}-clause}
\newcommand{\Bclause}[0]{\textsc{b}-clause}
\newcommand{\Aclauses}[0]{\textsc{a}-clauses}
\newcommand{\Bclauses}[0]{\textsc{b}-clauses}

\usepackage{hyperref}
\hypersetup{urlcolor=blue}
\hypersetup{colorlinks=true,urlcolor=blue}

\begin{document}
\title{
Where the Really Hard Quadratic Assignment Problems Are: the QAP-SAT instances
}
\titlerunning{Where the Really Hard QAP Are: the QAP-SAT instances}

\author{S\'ebastien Verel$^1$ \and Sarah L. Thomson$^2$ \and Omar Rifki$^1$}

\institute{
	 $^1$LISIC, Universit\'e du Littoral C\^ote d'Opale (ULCO), France \\
	\email{verel,omar.rifki@univ-littoral.fr}\\
  $^2$Edinburgh Napier University, Scotland, United Kingdom \\
	\email{S.Thomson4@napier.ac.uk}
}
\authorrunning{S. Verel \and S. L. Thomson \and O. Rifki}
\maketitle
\begin{abstract}
The Quadratic Assignment Problem (QAP) is one of the major domains in the field of evolutionary computation, and more widely in combinatorial optimization. This paper studies the phase transition of the QAP, which can be described as a dramatic change in the problem's computational complexity and satisfiability, within a narrow range of the problem parameters. To approach this phenomenon, we introduce a new QAP-SAT design of the initial problem based on submodularity to capture its difficulty with new features. This decomposition is studied experimentally using branch-and-bound and tabu search solvers. A phase transition parameter is then proposed. The critical parameter of phase transition satisfaction and that of the solving effort are shown to be highly correlated for tabu search, thus allowing the prediction of difficult instances. 

\end{abstract}

\section{Introduction}

The Quadratic Assignment Problem (QAP) has held major importance within evolutionary computation research for decades~\cite{taillard1991robust,ahuja2000greedy,benlic2015memetic,achary2021performance}. Given a matrix of flow between abstract objects, and a distance between positions, the goal of QAP~\cite{koopmans1957assignment} is to find an assignment of objects to positions in order to minimize the sum of costs, \textit{i.e.} the product of flow and distance, between all possible pairs of objects. The QAP is often considered one of the most difficult problems in the NP-hard class with many applications~\cite{loiola2007survey}. Notice that Traveling Salesperson Problem (TSP) is a special case of QAP with a dedicated flow matrix.
In order to understand and improve optimization algorithms design, a large number of QAP instances with relevant properties have been proposed in the literature (see more details in Section~\ref{sec:qap}).

Coming from statistical physics, the notion of a phase transition is also an important property in combinatorial optimization and for decision problems. Phase transition is a phenomenon related to the rapid change around a critical value of an order parameter of the probability that a random instance is satisfiable. From the seminal work on SAT problems~\cite{cheeseman1991really}, this property has been shown in a large number of decision and combinatorial problems~\cite{hartmann2006phase,biroli2002phase} (TSP, constraint problems, etc.).
The phase transition is also associated with problem difficulty which also changes around the point of the phase transition. Indeed, problem instances defined around the critical value of phase transition between feasible and unfeasible are often considered as the most challenging, and the most interesting instances to solve.
However, except for highly generic assignment problems~\cite{buhmann2017phase} different from classical QAP, or special cases of QAP such as TSP~\cite{gent1996tsp}, to our best knowledge no phase transition properties have been shown for QAP.
One objective of this work is to show a phase transition in the "pure" QAP problem. To show the phase transition phenomenon, we propose new QAP instances with tunable difficulty: the QAP-SAT, based on submodularity (clauses) similar to the notion of a clause in SAT problem.\\
\textcolor{white}{.}\quad \textit{Phase transition, and problem difficulty.}
\label{sec:phase_transiton}
Phase transition is a phenomenon that appears in randomly generated instances of intractable decision problems. 
It links the computational complexity to the satisfiability of the instances, such that a sudden change in the satisfiability happens in a narrow range of the instance parameters, summarized by an order parameter.
The passage from easily solvable satisfiable problem instances to easily checked unsatisfiable problem instances can be seen. 
At the transition between both regions, where the order parameter crosses the critical value, hard random instances can be found.
As this critical value has been shown to be independent of the solving algorithms, hard and easy instances can thus be located with precision, and subsequently used for fair benchmarking, algorithm selection, and configuration.
The first applications on NP-complete problems date back to \cite{cheeseman1991really}. 
From this point, phase transition has been observed on most of the famously known NP-complete problems, such as the satisfiability problem \cite{cheeseman1991really, gent1994sat, monasson2007introduction}, the traveling salesman problem \cite{cheeseman1991really, gent1996tsp}, the graph coloring problem \cite{cheeseman1991really}, the 0-1 knapsack problem \cite{yadav2018phase}, the minimum vertex cover problem \cite{weigt2000number}, and so on.\\
\textcolor{white}{.}\quad Another way of characterizing the inherent structure of the search space is fitness landscape analysis: this provides a number of features. These landscape metrics happen to be valuable for gaining insight into the performance of algorithms on a given problem instance, thus relating to problem difficulty. We discuss landscape analysis for QAP in Section 2.3.
For a general introduction to the topic of phase transition and problem difficulty, see \cite{leyton2014understanding, smith2012measuring}.
It should be also mentioned that phase transition on random combinatorial optimization problems can be seen through the prism of statistical physics of disordered systems \cite{hartmann2006phase, biroli2002phase}. Models such as the spin glass model are used to exhibit the change in behavior of the problem satisfiability, for example, \cite{monasson1999determining} for K-SAT.\\
\textcolor{white}{.}\quad The remaining of the paper is as follows.
Section 2 presents the QAP, its formulation, famous benchmark datasets, and its problem features.
Section 3 introduces the QAP-SAT decomposition.
The experimental study is detailed in Section 4. Section 5 concludes the paper.

\vspace{-0.2cm}
\section{Quadratic Assignment Problems}
\label{sec:qap}

\vspace{-0.2cm}
\subsection{Definition of QAP}

The Quadratic Assignment Problem (QAP)~\cite{koopmans1957assignment} is a minimization assignment problem where the search space ${\cal S}_n$ is the symmetric group of dimension $n$, \textit{i.e.} the set of permutations of size $n$. The QAP fitness function is defined as follows:
\vspace{-0.2cm}
$$
\forall \sigma \in {\cal S}_n, \ Q_{A,B}(\sigma) = \sum_{i=1}^n \sum_{j=1}^n A_{ij} B_{\sigma_i \sigma_j}
$$
where $A$, and $B$ are square matrices of real numbers of dimension $n \times n$. 
Usually, $A$ is called a flow matrix. $A_{ij}$ represents the flow (cost) between two abstract objects $i$, and $j$. $B$ matrix is called distance matrix. $B_{ij}$ represents the distance (cost) between positions $i$, and $j$. The objective function has a quadratic form: this is the sum for possible couples of objects $i$, and $j$ of the assignment cost defined as the product of flow by distance.

As such, usually $A_{ij}$, and $B_{ij}$ are positive real numbers. 
The matrix $B$ could, in fact, represent a distance matrix (triangular inequality is fulfilled), but this is not necessary. 
Likewise, the matrix $B$ could be symmetric, but this is not necessary in the general case.
Here, we will only consider that the self distance $B_{ii}$ is equal to $0$ for all $i \in [n]$. As a consequence, $A_{ii}$ can be considered as equal to $0$ for all $i \in [n]$.

\vspace{-0.18cm}
\subsection{QAP benchmark}

QAP has a lot of applications in real world~\cite{commander2005survey}.
As a consequence, a lot of benchmark instances have been proposed to understand problem difficulty, or to design more efficient optimization algorithms according to the features of QAP~\cite{loiola2007survey}.
The most well-known benchmark of QAP problems is the QAPLib~\cite{burkard1997qaplib}. QAPLib is a collection of instances with real-world problems usually of small size, and larger artificial ones generated with specific properties. 
The most used artificial instances are probably the two series of Taillard instances (\textit{Taia}, and \textit{Taib})~\cite{taillard1995comparison}. In the uniform instances (\textit{Taia}), the distance matrix is a Euclidean distance matrix between random points in a circle, and the flow matrix is random matrix with integer randomly selected between two bounds.
The real-like instances (\textit{Taib}) are inspired from some real world problems and mimic some of their properties. The distance matrix is also an Euclidean matrix but where the points are clustered, and the values of the flow matrix are exponentially distributed. Indeed, a lot of entries of the flow matrix are equal to zero.
It is well demonstrated that uniform instances are more challenging than real-like instances~\cite{verel2018sampling}.

Other instances have been proposed. The \textit{Taie} and \textit{Dre} series of instances were specifically designed to be difficult for metaheuristics~\cite{drezner2005recent}. Additionally, St\"utzle \textit{et al.} generated instances which vary two instance parameters systematically; these are related to flow dominance and sparsity~\cite{stutzle2004new}. A special case of QAP which is polynomially solveable~\cite{laurent2015quadratic} has been also proposed in order to test "black-box" algorithms.
Designing relevant QAP instances and understanding their properties has a lot of interest because doing so can facilitate the testing and improvement of optimization algorithms.

\vspace{-0.15cm}
\subsection{Features and Problem Difficulty in QAP}

Instances of the QAP were first characterised by the notion of flow dominance \cite{vollmann1966facilities}, which measures imbalance in the $A$ and $B$ matrices. A very high dominance value would be obtained if there is a substantial distance between comparatively few locations, or if there is a high degree of flow between only a few facilities. Measurement of sparsity for the two matrices has also been used in the literature \cite{smith2008towards}: this is the number of zero-entries as a proportion of the \(n^2\). 

 The QAP has served as one of the main testbeds for fitness landscape analysis of combinatorial spaces and several measures have been shown to be linked to search difficulty in some way. The correlation length --- which captures how far apart solutions with related fitness are --- and the fitness-distance correlation --- the connection between distance and fitness among local optima --- were related to the performance of memetic search algorithms for QAP \cite{merz2000fitness}. A set of measurements including some from information theory were computed from walks on QAP landscapes and used to aid in algorithm decisions \cite{pitzer2013automatic}. Another study focused considered whether landscape measures might be linked to the nature of the instance specification (distance and flow matrices) \cite{tayarani2015quadratic}, finding that autocorrelation and the size of plateaus were affected by the number of similar values in the matrices. The local optima space of QAP instances has also been studied \cite{daolio2010local}. 
 
 Fourier decomposition has been applied to the QAP: a branch-and-bound approach which operates in the Fourier space \cite{kondor2010fourier} has been proposed. Elementary landscape decomposition has also been leveraged \cite{chicano2010elementary}; this proved that the QAP, through the prism of the pairwise swap neighbourhood, can be represented as the combination of exactly three elemental landscapes. 

\section{Definition of QAP-SAT}

\subsection{Generic QAP-SAT}

The idea of QAP-SAT is to define the fitness function as a sum of easy QAP problems, called clauses, which depend only on a few variables. QAP-SAT follows the principle of the MAX-SAT problem, which is defined as the sum of the satisfaction of each clause, where clauses are easy low dimensional pseudo-boolean problems.
In QAP-SAT, we say that a clause is satisfied when a candidate solution of the problem reaches the lower bound of the clause.
Although there are similarities, a difference to the MAX-SAT space of functions is that the set of functions for QAPs is not a vector space~\cite{elorza2022characterizing}. In general, the sum of two QAP problems defined on ${\cal S}_n$ is not a QAP problem on ${\cal S}_n$. The QAP space has a bi-linear property: for any matrices $A, A^{\prime}, B, B^{\prime}$ of dimension $n \times n$,
$
Q_{A + A^{\prime}, B + B^{\prime}} = Q_{A, B} + Q_{A, B^{\prime}} + Q_{A^{\prime}, B} + Q_{A^{\prime}, B^{\prime}}
$.
The linear property between QAP problems is preserved when the same distance matrix B is shared. For any distance matrix $B$, for any positive integer $m$, and any square matrices $A_1, \ldots A_m$: 
$Q_{\sum_{\alpha=1}^{m} A_{\alpha}, B} = \sum_{\alpha=1}^{m} Q_{A_{\alpha}, B}$.
In the following, we define a clause for QAP by defining a clause both for flow matrix $A$, and distance matrix $B$.

\textit{\Aclause.}
A matrix $A$ of dimension $n \times n$ is an \textit{\Aclause}~of size $k>0$ when $\forall i \in [n]$ $A_{ii} = 0$, and it exists a subset $V_{A} \subset [n]$ of size $k$ such that:
$\forall (i,j) \in V_{A}^2$, $i \not= j$, $A_{ij} > 0$, and $\forall (i,j) \not\in V_A^2$, $A_{ij} = 0$.
The left two matrices in Figure~\ref{fig:exBclause} form an example of \Aclause.

\textit{QAP-SAT clause.}
A QAP problem $Q_{A,B}$ is a clause of size $k$ for the matrix $B$ iff $A$ is an \Aclause~of size $k$.

When the flow matrix $A$ is an \Aclause~of size $k$, 
the computation of the corresponding clause $Q_{A,B}$ is reduced to the sum of the $k(k-1)$ non-zero terms:
$
\forall \sigma \in {\cal S}_n, \ Q_{A,B}(\sigma) = \sum_{(i,j) \in V_A^2, i \not= j} A_{ij} B_{\sigma_i \sigma_j}
$
In this case, a lower bound $\text{lb}(Q_{A,B})$ of the $Q_{A,B}$ is: $\ell \sum_{(i,j) \in V_A^2, i \not= j} A_{ij}$ where $\ell = \min\{ B_{ij} : (i,j) \in [n]^2 \}$.
For example for the matrix $A_3$, this lower bound for $Q_{A_3,B}$ is equal to $10$.
A \Bclause~for the distance matrix is defined to ensure that the previous lower bound can be reached for a clause problem $Q_{A,B}$. In this work, without losing generality, the minimum non-zero value of the distance matrix is fixed to $1$. This value can be fixed to an arbitrary positive value. In this case, all values would be scaled to the selected non-zero minimum.

\textit{\Bclause.}
A matrix $B$ of dimension $n \times n$ is a \textit{\Bclause}~of size $k > 0$ when $\forall i \in [n]$ $B_{ii} = 0$, and there exists a set $V_{B} \subset [n]$ of size $k$ such that:
$\forall (i,j) \in V_{B}^2$, $i \not= j$, $B_{ij} = 1$, and $\forall (i,j) \not\in V_B^2$, $B_{ij} = \textsc{m}$ where $\textsc{m}>1$ is the largest possible distance of the problem.

As a consequence, when $A_k$ is an \Aclause~of size $k$, and $B_k$ is a \Bclause~of size $k$, the minimum of the clause $Q_{A_k, B_k}$ is the lower bound $\text{lb}(Q_{A_k, B_k})$.
More generally, a clause $Q_{A,B}$ is said \textit{satisfied} iff the minimum of $Q_{A,B}$ is equal to the lower bound $\text{lb}(Q_{A,B}) = \sum_{(i,j) \in V_A^2, i \not= j} A_{ij}$.
The QAP-SAT problem is an aggregation of \Aclauses~and \Bclauses. Let us define the aggregation principle of \Bclauses.
As for the Hadamard product, let us define the matrix $B \odot B^{\prime}$ by taking the minimum element by element: $\forall (i,j) \in [n]^2, (B \odot B^{\prime})_{ij} = \min \{ B_{ij}, B^{\prime}_{ij} \}$. Notice that $\odot$ is an associative operator.
We say that a distance matrix $B$ is composed of $m_1$ \Bclauses, with $m_1 > 0$, when it exists $m_1$ \Bclauses~$B_1, \ldots, B_{m_1}$, and a matrix $C$ with $\forall i \not= j$, $C_{ij} > 1$, and $\forall i, C_{ii}=0$, such that $B = B_1 \odot B_2 \odot \ldots \odot B_{m_1} \odot C$.

\textit{QAP-SAT.}
The QAP problem $Q_{A,B}$ is a \textit{QAP-SAT} problem with $m$ \Aclauses, and $m_1$ \Bclauses~when the matrix $B$ is composed of $m_1$ \Bclauses, and it exists $m$ clauses $Q_{A_1, B}, \ldots, Q_{A_m, B}$ for the same matrix $B$ such that $Q_{A,B}$ is the sum of those $m$ clauses: $Q_{A, B} = \sum_{\alpha=1}^{m} Q_{A_{\alpha}, B}$.
A QAP-SAT problem $Q_{A,B}$ is \textit{satisfied} when all clauses are satisfied, \textit{i.e.} when all clauses reach the lower bound:
$
\min Q_{A,B} = \sum_{\alpha=1}^m \text{lb}(Q_{A_\alpha, B})
$.

\begin{figure}
\centering
$$
A_3 =
\begin{bmatrix}
0 & 0 & 0 & 0 & 0 \\
0 & 0 & 1 & 0 & 2 \\
0 & 2 & 0 & 0 & 1 \\
0 & 0 & 0 & 0 & 0 \\
0 & 3 & 1 & 0 & 0 \\
\end{bmatrix}
A^{(3)} =
\begin{bmatrix}
0 & 1 & 2 \\
2 & 0 & 1 \\
3 & 1 & 0 \\
\end{bmatrix}~
B_3 =
\begin{bmatrix}
0 & 1 & \textsc{m} & \textsc{m} & 1 \\
1 & 0 & \textsc{m} & \textsc{m} & 1 \\
\textsc{m} & \textsc{m} & 0 & \textsc{m} & \textsc{m} \\
\textsc{m} & \textsc{m} & \textsc{m} & 0 & \textsc{m} \\
1 & 1 & \textsc{m} & \textsc{m} & 0 \\
\end{bmatrix}
B^{(3)} =
\begin{bmatrix}
0 & 1 & 1 \\
1 & 0 & 1 \\
1 & 1 & 0 \\
\end{bmatrix}~
B =
\begin{bmatrix}
0 & 1 & 2 & 4 & 1 \\
1 & 0 & 4 & 2 & 1 \\
3 & 8 & 0 & 5 & 3 \\
3 & 6 & 7 & 0 & 2 \\
1 & 1 & 2 & 5 & 0 \\
\end{bmatrix}
$$

\caption{For problem dimension $n=5$, examples of \Aclause~and \Bclause~of size $k=3$ with $V_A = \{ 2, 3, 5 \}$, and $V_B = \{ 1, 2, 5 \}$. $A^{(3)}$, and $B^{(3)}$ are sub-matrix with variables of $V_A$, and $V_B$. 
Distance matrix $B$ composed of $m_1=1$ \Bclause~complementary to matrix $B_3$.} \label{fig:exAclause}
\label{fig:exBclause}
\end{figure}

\subsection{Random QAP-k-SAT}

When all clauses (\Aclauses~and \Bclauses) have the same size $k>0$, the QAP-SAT is denoted QAP-k-SAT.
The random QAP-k-SAT is designed with the same principle of the classical random k-SAT problem. The $k$ variables of each clause are randomly and independently selected.
The random QAP-k-SAT is defined by 4 basic parameters. The problem size $n$, the size of the clause $k$, the number $m$ of \Aclauses, and the number $m_1$ of \Bclauses.

For each \Aclause, $k$ different variables are randomly selected. In this work, the size of clauses is set to $k=3$. It is possible to create any random sub-matrix of size $k$ for \Aclause. However, in this work for simplicity (same lower bound for example), the \Aclauses~ are based on the same sub-matrix $A^{(3)}$. Only the order of variables is randomly swapped.
For each \Bclause, $k$ different variables are also randomly selected. Notice that the minimal \Bclause~is symmetric (see matrix $B^{(3)}$), no need to swap randomly the variables.
Several choices can be made to create the complementary matrix $C$ of distances which have non-minimal values. Of course a basic choice would be to create random integer numbers between 2 and a maximal value. However, in this work, we prefer a slightly more sophisticated choice. All values of $C$ are integer values higher or equal than 2. They are selected in order to have geometric distribution of values.
Let $n_d$ be the number of values equal to $d$ in the matrix $B$: $n_d = \sharp \{ B_{ij} = d ~:~ (i,j) \in [n]^2 \}$, and $p_d$ the respective proportion in the matrix: $p_d = n_d / (n (n-1))$. For $d > 1$, the proportion $p_d$ is set in order to have approximately $p_d = p_1^d$ (to the precision of integer values for $n_d$). Indeed, we set $n_d = \max \{1, \lceil p_1( n (n-1) - \sum_{\delta = 1}^{d - 1} n_\delta ) \rceil \}$. Then, the position of values is randomly distributed in $B$ on available positions in the matrix. As a consequence, the matrix $B$ is not necessary symmetric.
Figure~\ref{fig:exBclause} is an example of distance matrix.
Python code for the generator of QAP-SAT instances, and the instances used in this work are available on the repository: \url{https://gitlab.com/verel/qap-sat}.

\section{Experimental analysis}

\subsection{Experimental design}

\paragraph{Instance Generation}

\begin{table}[t]
    \centering
    \caption{Parameter settings of the random QAP-3-SAT instances.
    The sets corresponds to parameters for dimensions $n=18,19$.}
    \begin{tabular}{|c|l|l|}
\hline
Name & Description & Values \\
\hline
$n$   & Problem dimension & $\{8, 9, \ldots, 17 \} \ \ \ \ \{ 18, 19 \}$ \\
\hline
$k$   & Size of clause & 3 \\
\hline
$m_1$ & Number of \Bclauses~(distance matrix) & \small{$\{ 3, 6, 9, \ldots, 27 \} \ \{ 3, 9, 15, \ldots, 57 \}$}\\
\hline
$m$   & Number of \Aclauses~(flow matrix) & $\{ 1, 2, 3,\ldots, 40 \} \ \{ 1, 3, 9, 15, \ldots, 57, 63 \} $\\
\hline
    \end{tabular}
    \label{tab:param}
\end{table}

In this work, we generate small and medium size instances of QAP-SAT from the dimension $n=8$ to the dimension $n=19$. The instance parameters are given in Table~\ref{tab:param} to generate a factorial design. $50$ instances for each parameter triplet $(n, m_1, m)$ have been generated. Thus, for dimension $n$ lower than 17, $18,000$ instances are generated for problem dimension. To reduce the computation time, for problem dimensions $18$ and $19$, we reduce the number of instances to $6,000$. In total, $192,000$ instances have been analyzed.
Small size instances can be fully enumerated until dimension $n \leq 13$, then we use a branch and bound algorithm to compute global minima (see next paragraph).

\textit{Branch and Bound algorithm.}
An exact algorithm is required to find the global minimum of each instance, in particular for medium size instances with dimension larger than 14. First, we test a classical Cplex algorithm with a standard linear transformation of QAP, provided by~\cite{silva2021quadratic}\footnote{\url{https://github.com/afcsilva/PMITS-for-QAPVar}}, but the computation time is too long for our experimental scenario. For instance, the average computation time for the small dimension $n=10$, across all $m$ and $m_1$ values is equal to $50.6$ seconds. Indeed, this first experiment -- not detailed in this paper for the sake of brevity -- is an indication that the QAP-SAT instances can be difficult to solve.

Several Branch and Bound (B\&B) algorithms have been proposed to solve QAP from the seminal works based on Gilmore-Lawler bound~\cite{gilmore1962optimal}. In this work, we use a recent and efficient B\&B algorithm proposed by Fujii \textit{et al.}~\cite{fujii2021solving} for which the MATLAB code is available\footnote{\scriptsize{\url{https://sites.google.com/site/masakazukojima1/softwares-developed/newtbracket?pli=1}}}. The algorithm is based on the Lagrangian doubly non-negative relaxation and Newton-bracketing. Please refer to the original article for details. The goal of this work is not to compare the efficiency of B\&B algorithms for solving QAP-SAT instances, but to find the global minimum and estimate the computation time to find it as a possible measure of difficulty.

\textit{Robust Taboo Search.}
We use Taillard's implementation in C of his robust taboo search (ROTS) algorithm for the QAP \cite{taillard1991robust}\footnote{\url{http://mistic.heig-vd.ch/taillard/codes.dir/tabou_qap2.c}}; this is considered a competitive metaheuristic for the domain. The neighbourhood is a random pairwise swap in the permutation, and parameters are kept as those provided in the code: tabu duration is \(8n\); aspiration is set at \(5n^2\); and runs terminate when the global optimum is found or after 1000 solutions have been visited by the search. The global optimum has been computed for all considered instances; proportional success rate is therefore computed as a metric of performance. We compute the mean for this metric over 30 runs per instance. 

\subsection{Phase transition of satisfaction probability}
\label{sec:phase}

\begin{figure}[ht!]
    \centering
\includegraphics[width=\textwidth]{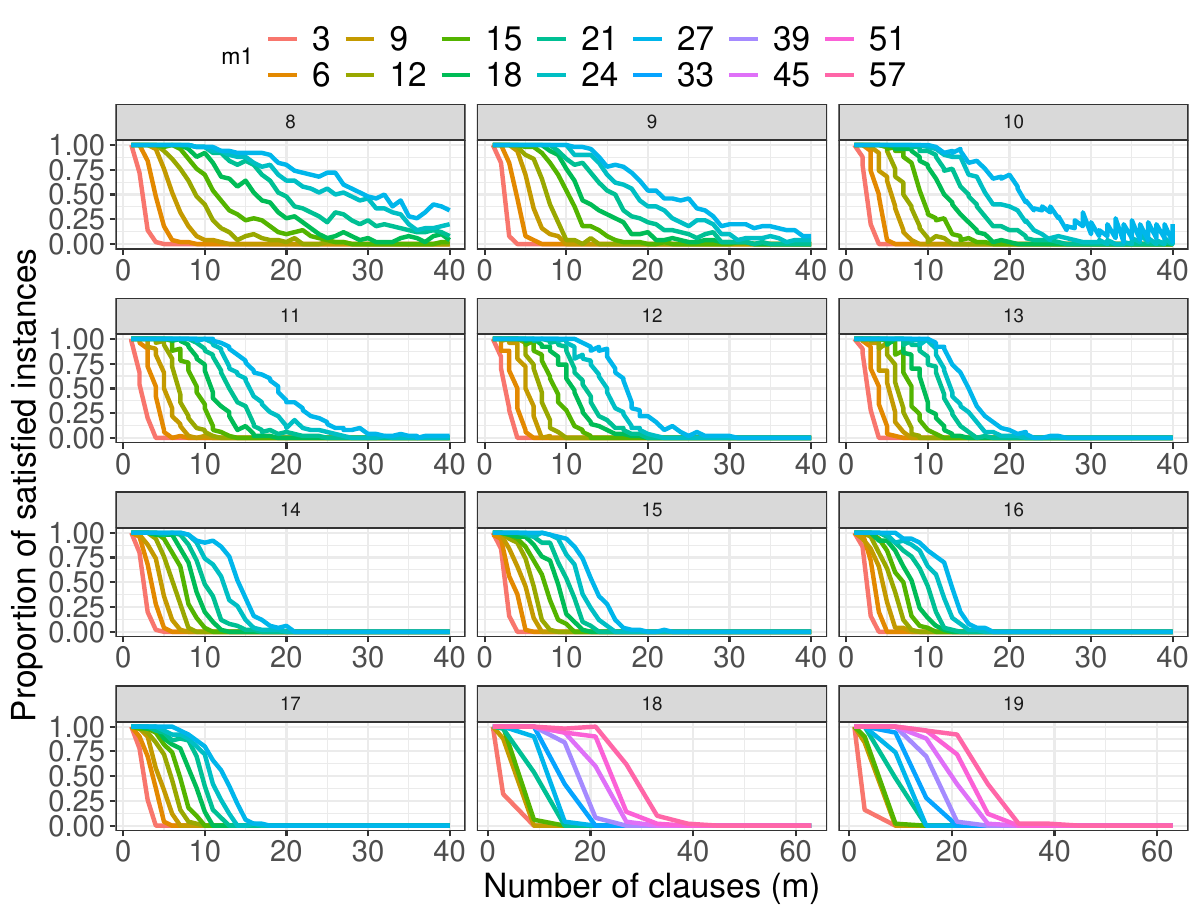}
\caption{Proportion of satisfied instances, for which the minimal values is the lower bound, according to the number of \Aclauses~$m$. Facet: problem dimension $n$.}
\label{fig:solved}
\end{figure}

A phase transition in combinatorial optimization is characterized by a rapid change between satisfied and non satisfied instances according to a phase parameter.
Figure~\ref{fig:solved} shows the proportion of satisfied instances for which the minimum is the lower bound \textit{i.e.} all clauses of the problem are satisfied with the minimal possible value.
As expected, when the number \Aclauses~$m$ --- the number of clauses for the flow matrix --- is small, the probability to have an instance satisfied is nearly equal to $1$. This probability drops very quickly around a critical value denoted $m_c$. When the number $m$ of \Aclauses~is much larger than the number of \Bclauses~$m_1$ --- the number of clauses for the distance matrix --- none of the instances are satisfied. The only exception is for small dimension $n=8,9$ for which a large number of \Bclauses~$m_1$ gives a full distance matrix of $1$, and then all solutions are global optima.
In general, this curve seems to describe a sigmoidal shape dropping quickly around a critical value $m_c$.
Indeed, for a given problem dimension, the critical value $m_c$ increases with the number $m_1$ of \Bclauses. For instance, for problem dimension $n=10$, $m_c$ is around $8$ when $m_1 = 9$, and around $15$ when $m_1=21$.
But as problem dimension increases, the variation of $m_c$ according to $m_1$ is smaller. More details are given in the next section~\ref{sec:parameter}.

\vspace{-0.5cm}

\vspace{-0.6cm}
\subsection{Estimation of the phase transition parameter}
\label{sec:parameter}

The proportion of satisfied instances (see Fig.~\ref{fig:solved}) seems to follow a logistic function as a function of the number of \Aclauses~$m$. 
If it is true, it is possible to estimate the parameters of the logistic model, and compute the center of symmetry which correspond to inflection point of the logistic model. This center is the critical value of $m$ at the phase transition.
First, to estimate the parameters of the logistic regression, we use logit transform: if $p$ follows a logistic model, then $\text{logit}(p)= \log(\frac{p}{1 - p})$ follows a linear model, and inversely. If the regression model is $\text{logit}(p) = \beta_0 + \beta_1 m$, the abscissa of the center of the logistic model is $m_c = - \beta_0 / \beta_1$.

\begin{figure}[ht!]
    \centering
    \includegraphics[width=0.49\textwidth]{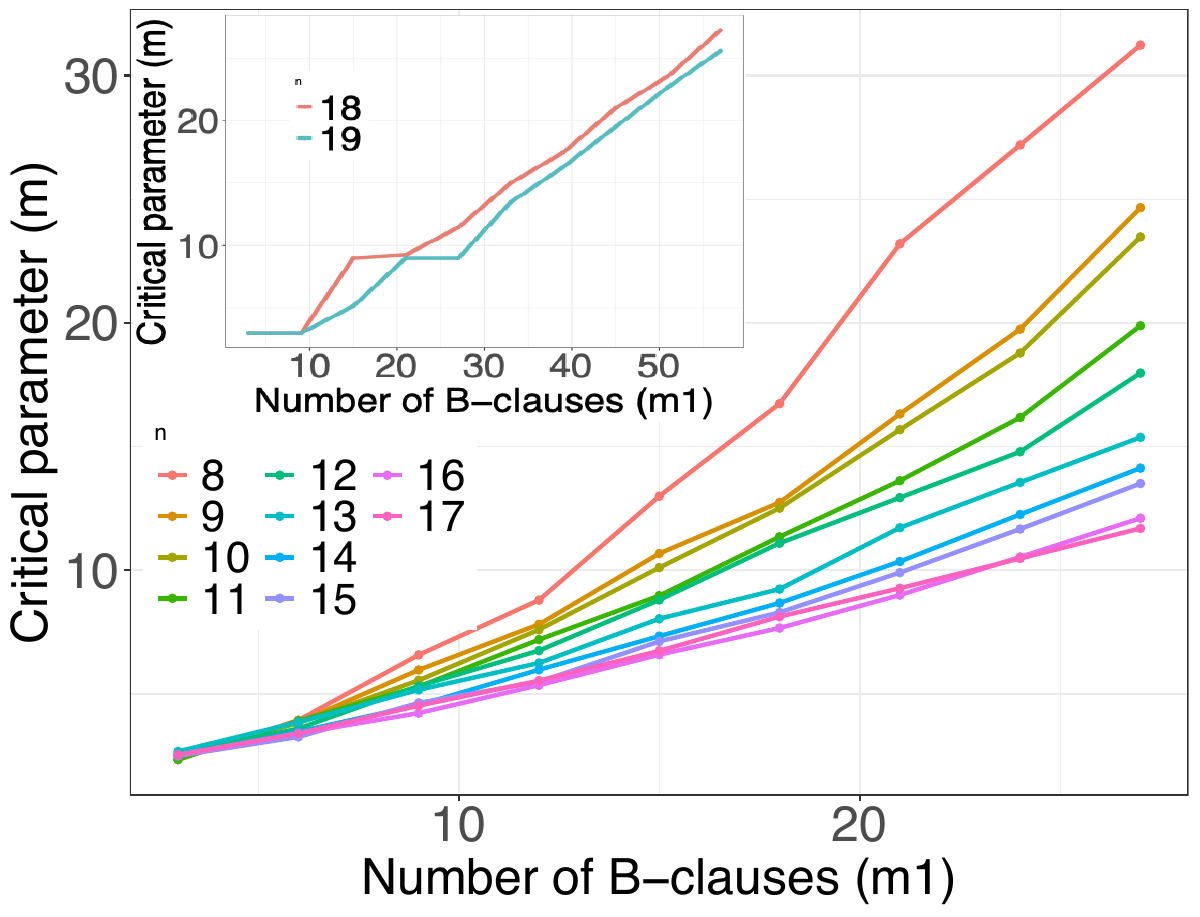}
    \includegraphics[width=0.49\textwidth]{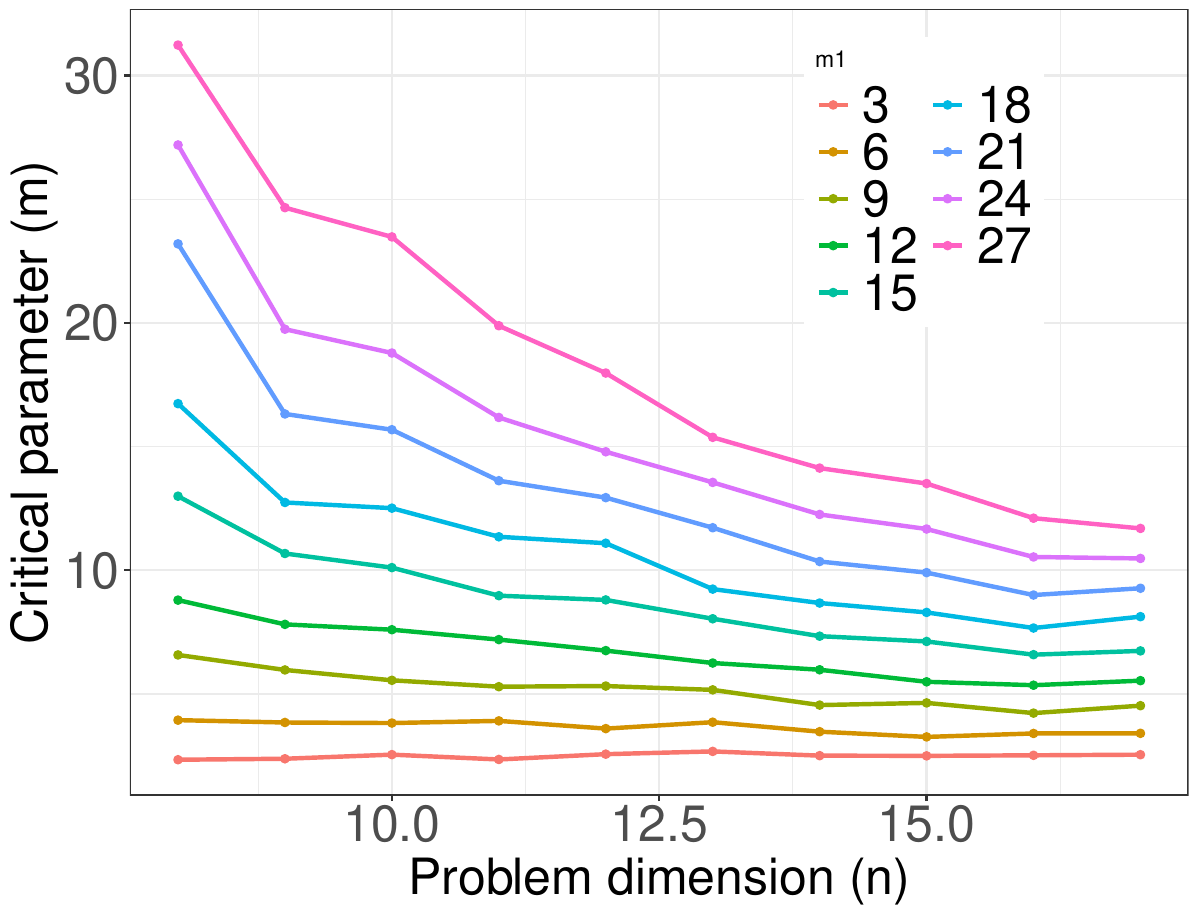}
    \caption{Critical parameter $m_c$ according to the number of \Bclauses~$m_1$ (left), and problem dimension $n$ (right).}
    \label{fig:mcritical}
\end{figure}

The critical value $m_c$ of the logistic model is estimated by the regression of the logit model.
Figure~\ref{fig:mcritical} shows the critical value $m_c$ of the number of clauses according to the number of minimal \Bclauses, and the problem dimension $n$. Those critical parameters have been estimated using the logit regression.
The $R^2$ values of the regression, which is not presented here to save space, are high. On the 110 possible instance parameters, the average $R^2$ is $0.9417$, and 98 instances have $R^2$ above $0.9$. The worst values of $R^2$ (minimum is $0.801$) are obtained for the smallest problem dimensions where the probability of satisfaction does not reach exactly $0$ for large $m$.

Except for small values of $n=8$ and $9$, the critical value $m_c$ as a function of \Bclauses~$m_1$ tends to be linear.
Table~\ref{tab:reg_mc_n} gives the estimated parameter values for a linear model $m_c = \beta_0 + \beta_1 m_1 + \epsilon$. For $n$ higher than 11, the adjusted $R^2$ regression quality is high: larger than $0.98$. The origin ordinate of the regression is close to 0, and the slope of the regression decreases with problem dimension $n$. At first, this suggests a linear dependency between critical value $m_c$ and $m_1$ when $n$ is "large" compared to $m_1$. Notice that the slope of the linear regression decreases with problem dimension $n$.
From the right side of Figure~\ref{fig:mcritical}, for a given value $m_1$, the critical value $m_c$ seems not always depend linearly on problem dimension $n$. Only for values small of \Bclauses~$m_1 \leq 9$, the critical value seems to be independent of problem dimension. 

\begin{table}[h]
\caption{Estimated parameter values of the regression model $m_c = \beta_0 + \beta_1 m_1 + \epsilon$.}\label{tab:reg_mc_n}

    \centering
\begin{tabular}{|r|l|l|l|}
\hline
$n$ & origin $\beta_0$ & slope $\beta_1$ & adj. $R^2$ \\ 
\hline
8 & -4.09305 & 1.258099 & 0.9702023 \\
9 & -1.960525 & 0.9022615 & 0.9679433 \\
10 & -1.290005 & 0.8273129 & 0.9797667 \\
11 & -0.5588647 & 0.7006147 & 0.9843299 \\
12 & -0.01016369 & 0.6367813 & 0.9872014 \\
13 & ~0.7395489 & 0.5280727 & 0.9923386 \\
\hline
\end{tabular}
~~~~~\begin{tabular}{|r|l|l|l|}
\hline
$n$ & origin $\beta_0$ & slope $\beta_1$ & adj. $R^2$ \\ 
\hline
14 & 0.4362908 & 0.4840685 & 0.9877075 \\
15 & 0.4972105 & 0.458738 & 0.9861831 \\
16 & 0.8578043 & 0.3977207 & 0.988515 \\
17 & 1.104765 & 0.3881437 & 0.9974293 \\
18 & 0.3636075 & 0.454062 & 0.9813189 \\
19 & -0.6242743 & 0.4427259 & 0.97458 \\
\hline
\end{tabular}
\end{table}

\begin{figure*}[ht!]
    \centering
    \includegraphics[width=0.5\textwidth]{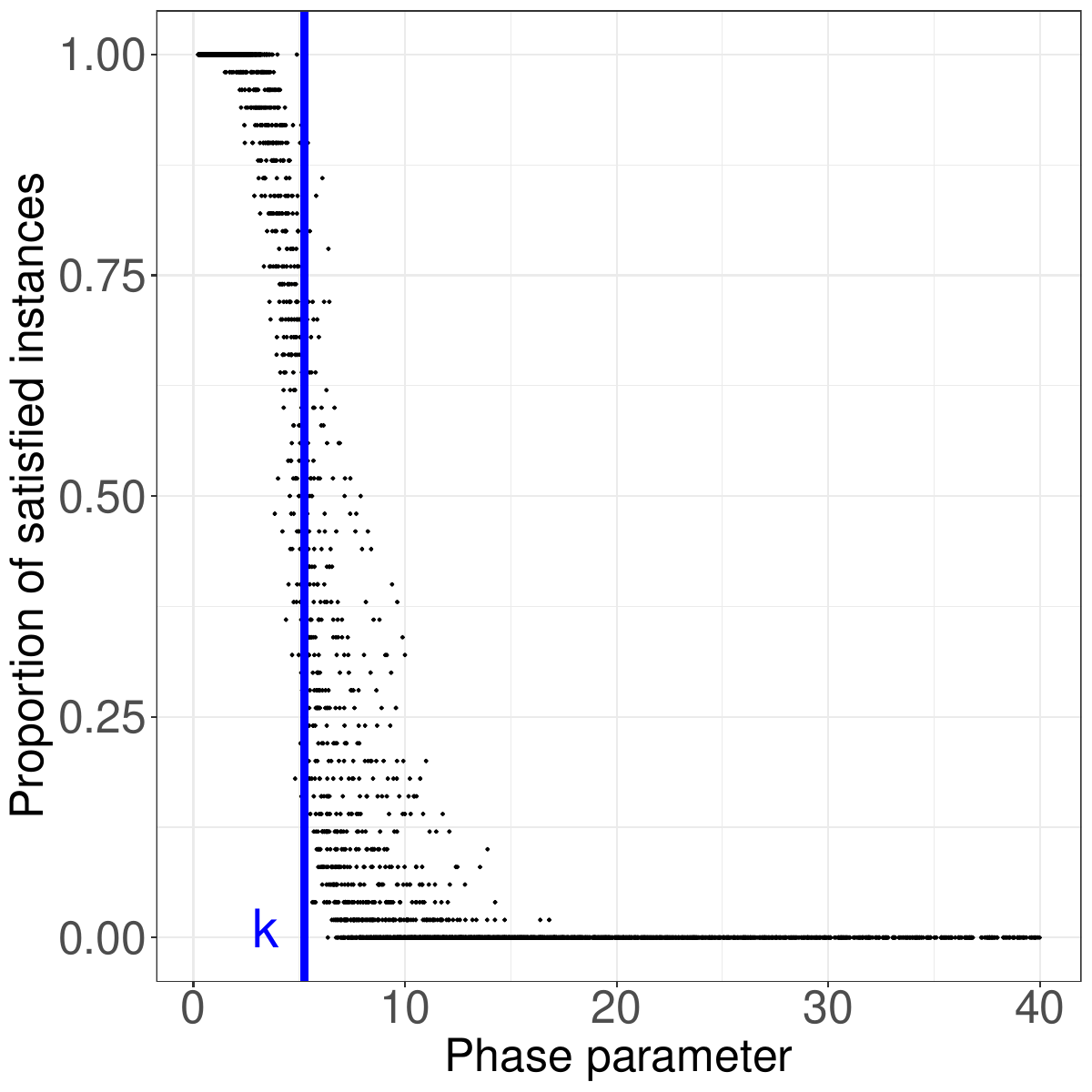}
    \caption{Probability of satisfied instances according to the phase transition parameter $\frac{m}{n^{\alpha_1} m^{\alpha_2}_1}$ across all instances. $k=\exp(1.65453) \approx 5.23$.}
    \label{fig:satisfied_param}
\end{figure*}

From the first observations, we can try to explain the critical value $m_c$ as a function of both $m_1$, and $n$: close to linear model as a function of $m_1$, but with a slope decreasing slowly with problem dimension. Moreover, from inspiration of SAT problem for which the phase transition parameter is the ratio between number of clauses and problem dimension, we run the following regression model:
$$
m_c = k  n^{\alpha_1} m_1^{\alpha_2} + \epsilon 
$$
where $k$ is a real constant value, and $\alpha_1$, $\alpha_2$ are exponents for variables $n$, and $m_1$, and $\epsilon$ is the noise of the regression model.
The parameters of the model can be estimated with a multi-linear regression on the logarithm of $m_c$: $\log(m_c) = \log(k) + \alpha_1 \log(n) + \alpha_2 \log(m_1)$.
We estimate the parameters using the values of $\log(m_c)$ for problem dimension below $17$. The adjusted $R^2$ of this regression is $R^2 = 0.947$ which gives an $R^2$ coefficient on the value $m_c$ (without log transformation) of $R^2 = 0.898$. The parameters of the regression are: $\alpha_1 = -0.75999$, and $\alpha_2 = 0.90365$, and $\log(k)=1.65453$. As expected from the previous linear regression, the exponent $\alpha_2$ for $m_1$ is close to the value $1$, and the exponent $\alpha_1$ for $n$ is negative between $-1/\sqrt{n}$, and $-1/n$ approximately. As $n$ increases, the slope of the linear relation between $m_c$, and $m_1$ decreases. When we test this model on instances with $n \geq 18$, the $R^2$ is higher, equal to $0.923$, which corroborates to the robustness of the model even if more data or a theoretical proof could help to validate it further.
So, we hypothesise that $\frac{m}{n^{\alpha_1} m_1^{\alpha_2}}$ is the phase parameter of QAP-3-SAT. To check this hypothesis, Figure~\ref{fig:satisfied_param} shows the proportion of satisfied instances as a function of this phase parameter. For all problem dimensions, and numbers of \Bclauses, the probability to have satisfied instances drops very quickly around the same value $k$.

\subsection{Performance of optimization algorithms}

\begin{figure}[ht]
    \centering
    \includegraphics[width=0.99\textwidth]{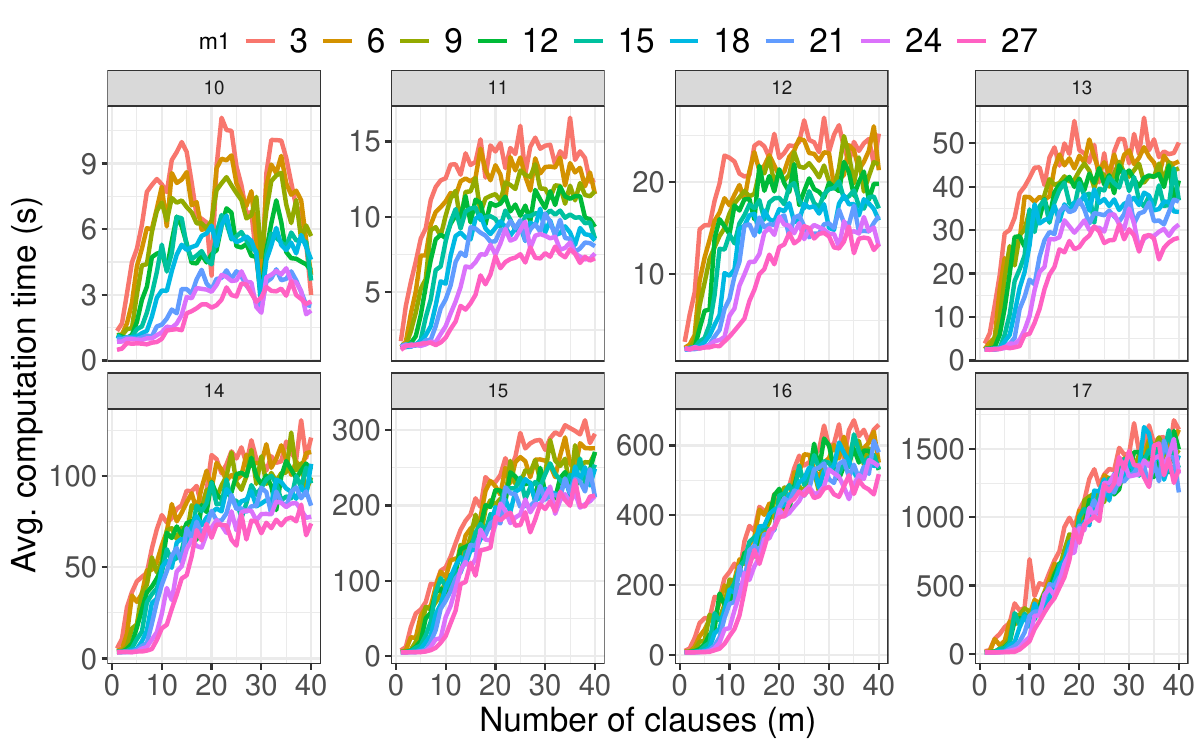}
    \caption{Average computation time of B\&B algorithm to find the global minimum. Problem dimensions $n$ = 10-17.}
    \label{fig:small_bb_time}
\end{figure}

\paragraph{Branch and Bound.} 
In the previous section, the Branch and Bound (B\&B) algorithm was used to find the global minimum. In this section, it is used to estimate the computational effort to find the global minimum. Notice that the B\&B algorithm is used in an optimization scenario to find and to prove global minima, and not in the decision scenario to decide only if a QAP-SAT instance is satisfied.
The Figure~\ref{fig:small_bb_time} shows the average computation time (in seconds) of the B\&B algorithm across instances sharing the same parameters. For fair comparison, the algorithm is run on the same computer configuration (Dell HPE DL385 with 512Go RAM, 2 processors AMD-EPYC Milan with 256M cache, and 48 cores at frequency 2.3Ghz each) for all instances.
The computation time increases by a larger factor when the problem dimension increases. For a given problem dimension $n$, the computation time drops from few seconds for instances with low-$m$, to high value after a threshold value $m$. Indeed, the computation time seems to follow a sigmoidal shape with the number of clause $m$. Around a critical value of $m$, the computation time increases fast.

\begin{figure}[ht]
    \centering
    \includegraphics[width=0.45\textwidth]{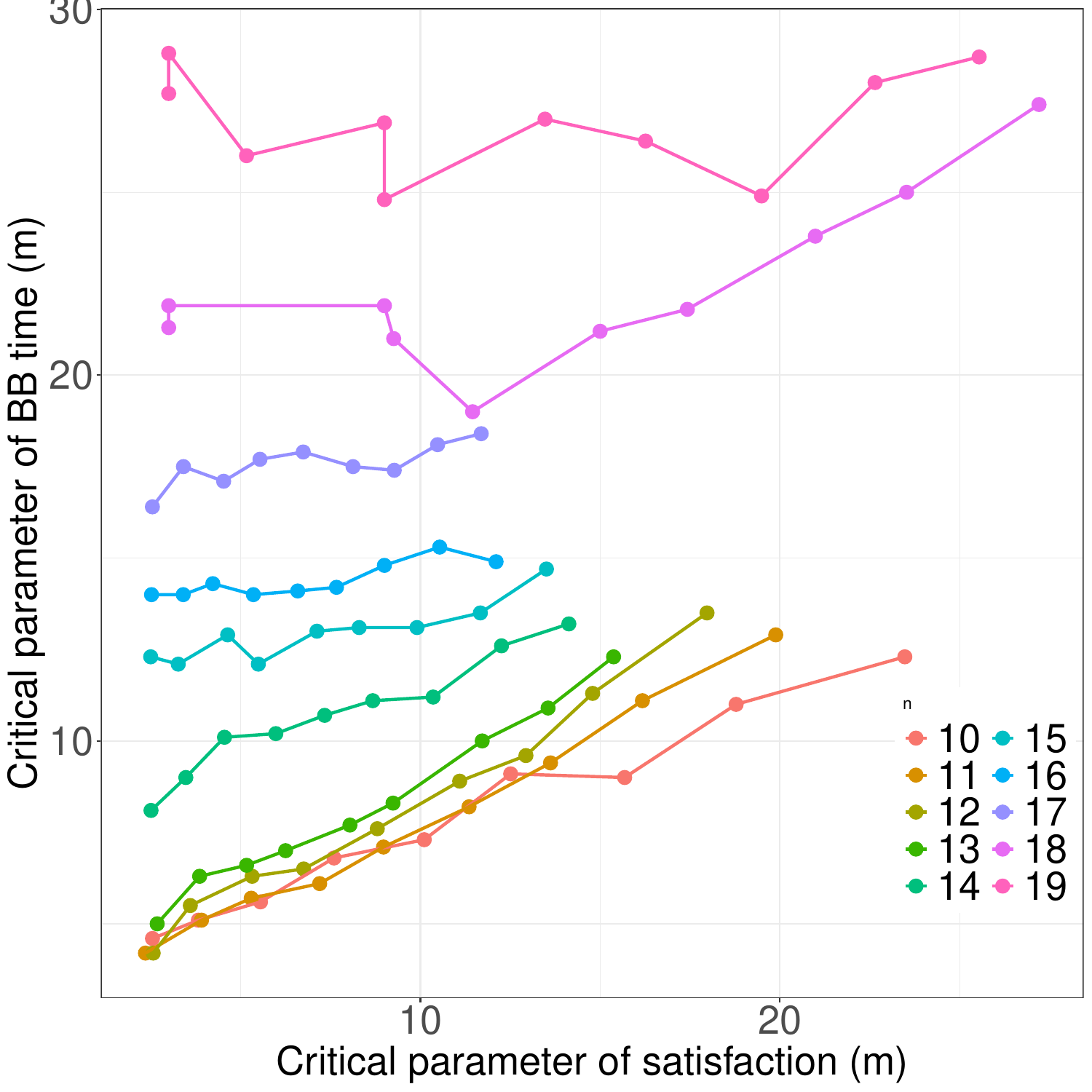}
    \includegraphics[width=0.45\textwidth]{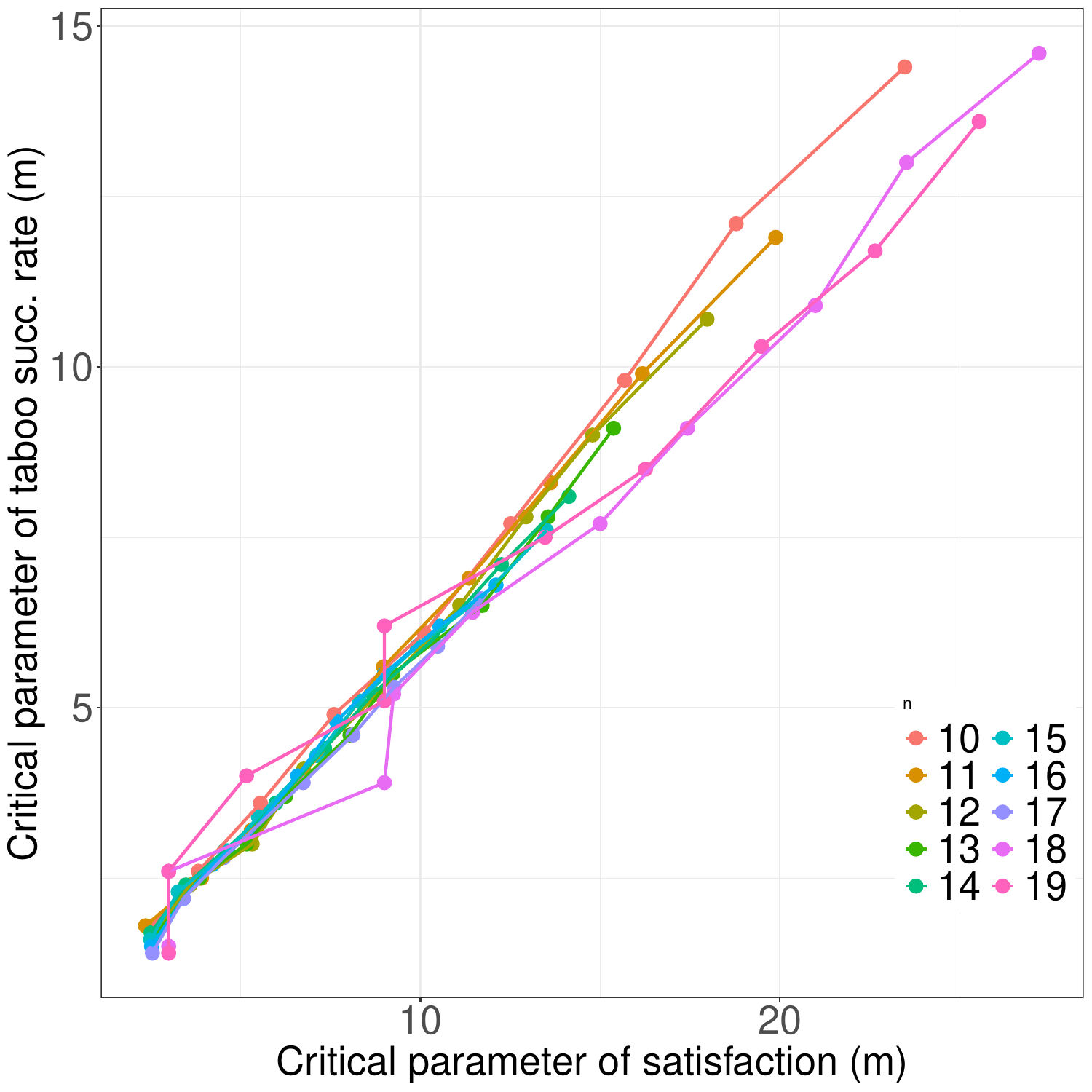}

    \caption{Relation between critical parameter $m_c$ of satisfaction transition, and critical parameter $m_t$ for B\&B average computation time (left), for taboo search success rate (right)}
    \label{fig:param_phase_algo}
\end{figure}

We analyze the relation between the critical parameter $m_c$ of satisfaction phase transition and the potential critical parameter of B\&B computation time.
First, we compute a regression using sigmoid function of average computation time: $t(m) = \frac{L}{1 + e^{- r (m - m_t)}}$ where $L$ is the maximal value, $r$ the rate of increases, and $m_t$ the inflection of the sigmoid \textit{i.e.} the critical parameter. Contrary to the previous section, due to noise of the computation time logit transformation can not be used to estimate the parameters. As we know the range of parameters, we use a basic grid search to estimate the regression parameters minimizing the mean square error.
Except for $n=10$ for which the variance of computation time is high and not stable, the $R^2$ coefficients of the regression are high --- over $0.925$ --- with a median equal to $0.969$. This result tends to show that the computation time of B\&B follows a sigmoidal shape like in the phase transition.
Figure~\ref{fig:param_phase_algo} (left) shows the relation between critical parameter $m_c$ of the satisfaction phase transition and the critical parameter of B\&B computation time.
For a given problem dimension, the relation is nearly linear. The average computation time depends on the phase transition parameter. However, the variance of the computation time critical parameter $m_t$ depends also on the problem dimension. For the same critical parameter $m_c$, the value of $m_t$ increases with problem dimension, and the range of variation with $m_c$ gets smaller as problem dimension increases. Indeed, we notice that the average of the maximum time $L$ across $m_1$ value for a given problem dimension $n$ is approximately given by $\gamma(2.043 + 0.476 (n - 8))$. To prove that a candidate solution is a global minimum, B\&B needs to cross a large part of the search anyway.

\vspace{-0.3cm}
\begin{figure}[h]
    \centering
    \includegraphics[width=0.99\textwidth]{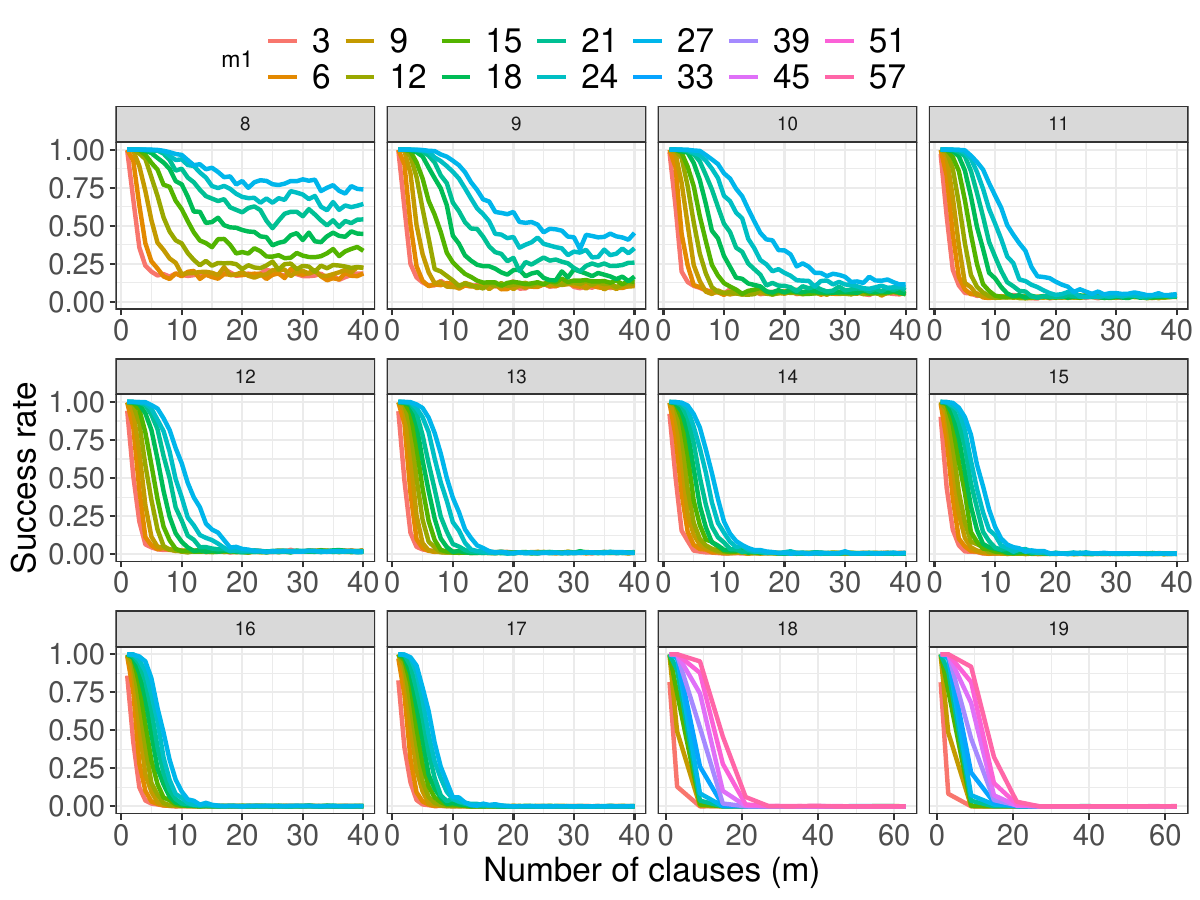}
    \caption{Success rate with varying number of clauses, $M$, for Robust Taboo Search. Problem dimensions $n$ = 8-19. }
    \label{fig:tabu-upto16-m}
\end{figure}

\vspace{-0.5cm}
\textit{Robust Taboo Search.} Figure \ref{fig:tabu-upto16-m} presents the success rate of ROTS for all considered instances. 
There is a very high negative correlation between success rate, and number of evaluations to reach global minimum ($\rho=-0.9999$). So only the success rate is analyzed in the following.
In the Figure, instances are split into facets according to problem dimension, $n$, and split by $m_1$ as individual lines. The horizontal axis is number of clauses in the instance, $m$. It follows that a single line in one of the plots represents, for all instances of the specified size and $m_1$, the mean ROTS success rate. 

Notice from Figure \ref{fig:tabu-upto16-m} that the success rate decreases with increasing $m$ for all problem dimensions. This effect is much stronger for larger problem sizes, however. From surveying individual facets it can be observed that lower values of $m_1$ are associated with lower success rates, although the disparities between low-$m_1$ and high-$m_1$ instances becomes substantially less for higher problem dimensions (compare, for example, the facet for problem size 8 with that of size 16). For low values of $m_1$, there is a steep decrease in success rate at approximately $m=5$. For larger $m_1$, the drop in success rate happens at a larger $m$, and for smaller problem sizes there is not a dramatic vertical drop for them. For problem sizes at 14 or greater, however, instances with all values of $m_1$ experience a steep drop in success rate between $m$ being 5 and approximately 21.

We also analyze the relation between the critical parameter $m_c$ of satisfaction phase transition, and the success rate of tabu search. Like for B\&B we obtain the estimate of a sigmoid model for success rate as a function of the number of clauses $m$ using a basic grid search minimizing mean square error. Except for very small values of \Bclauses~$m_1$, $R^2$ coefficients are high --- always over $0.9436$ with median equals to $0.9972$.
In contrast to B\&B, the correlation with the critical parameter of satisfaction phase transition is very high: $\rho = 0.9904$. The $R^2$ of the linear regression is $R^2 = 0.9811$. The slope of the linear relation is $0.542$.
This result shows that the difficulty to solve a QAP-SAT instance for tabu search is highly dependent on the position of the instance against the phase transition parameter. Recall that the tabu search stops when the global minimum is found; maybe the B\&B needs some additional time (which could depend on the problem dimension) to prove that the solution is the global minimum.

\section{Discussion and future work}

\vspace{-0.15cm}
In this work, QAP-SAT instances are designed using a submodularity principle by defining low-dimensional easy problems called "clauses" to solve jointly like in SAT problems. Each flow and distance matrix is composed of basic sub-matrices, and the agreement between sub-matrices in the flow and distance matrices sharply tunes the difficulty of the instance.
Although limited to medium size instances with problem dimension up to $19$, the large experimental analysis shows that the QAP-SAT problem shows a phase transition according to number of clauses in the matrices and the problem dimension. Supported by those first experiments, we suggest an order phase parameter that explains the phase transition. Moreover, the problem performance of the Branch \& Bound algorithm and robust taboo search are correlated with phase transition.

This initial work raises many open questions.
First, it is now possible to test optimization algorithms using QAP-SAT instances as a benchmark, including large size instances, and compare QAP-SAT against existing QAP benchmarks to stress the difference between them. Moreover, the modular design of QAP-SAT with clauses motivates the decomposition of existing real-world QAP instances into easy sub-problems. Additionally, it could inspire the definition of new properties and metrics on each matrix, and also jointly between matrices.
More broadly, it would be relevant to analyse the fitness landscape to understand the phase transition and problem difficulty in QAP-SAT.
From the theoretical side, a proof using existing mathematical techniques is required to support the proposed phase transition parameter based on experiments.
In this work, sub-problems of dimension 3 (QAP-3-SAT) have been studied using a specific shape for \Aclauses, and \Bclauses, but future works could extend the analysis to a broader class of clauses, or to relax the satisfiability condition of a clause, and the impact on phase transition parameters. Such extended analysis is likely to be a new field of questions for QAP-related structural properties and problem difficulty.


\bibliographystyle{splncs04}

\end{document}